\documentclass[final, twocolumn,5p,times, authoryear]{elsarticle}
\usepackage{amsmath}
\usepackage{booktabs}
\usepackage{url}
\usepackage{algorithm}
\usepackage{amsmath}
\usepackage{algpseudocode}
\usepackage{amsmath}
\usepackage{array}
\usepackage[colorlinks=False]{hyperref}
\usepackage[round]{natbib}
\usepackage{lineno}

\usepackage{amssymb}

\usepackage{orcidlink}
\begin{document}
\begin{frontmatter}



\title{SinkSAM-Net: Knowledge-Driven Self-Supervised Sinkhole Segmentation Using Topographic Priors and Segment Anything Model}

\tnotetext[t1]{This manuscript has been published in ISPRS Journal of Photogrammetry and Remote Sensing (2025). https://doi.org/10.1016/j.isprsjprs.2025.06.035}


\author[inst1]{Osher Rafaeli\corref{cor1}\raisebox{0.5ex}{\orcidlink{0000-0002-7097-7568}}}
\ead{osherr@post.bgu.ac.il}
\cortext[cor1]{Corresponding author.}

\author[inst1,inst2]{Tal Svoray\raisebox{0.5ex}{\orcidlink{0000-0003-2243-8532}}}

\author[inst1]{Ariel Nahlieli\raisebox{0.5ex}{\orcidlink{0009-0001-0633-1842}}}

\affiliation[inst1]{
    organization={Department of Environmental, Geoinformatics and Urban Planning Sciences, 
    Ben-Gurion University of the Negev},
    country={Israel}
}

\affiliation[inst2]{
    organization={Department of Psychology, 
    Ben-Gurion University of the Negev},
    country={Israel}
}

\begin{abstract}
Soil sinkholes significantly influence soil degradation, infrastructure vulnerability, and landscape evolution. However, their irregular shapes, combined with interference from shadows and vegetation, make it challenging to accurately quantify their properties using remotely sensed data. In addition, manual annotation can be laborious and costly. In this study, we introduce a novel self-supervised framework for sinkhole segmentation, termed SinkSAM-Net, which integrates traditional topographic computations of closed depressions with an iterative, geometry-aware, prompt-based Segment Anything Model (SAM). We generate high-quality pseudo-labels through pixel-level refinement of sinkhole boundaries by integrating monocular depth information with random prompts augmentation technique named coordinate-wise bounding box jittering (CWBJ). These pseudo-labels iteratively enhance a lightweight EfficientNetV2-UNet target model, ultimately transferring knowledge to a prompt-free, low-parameter, and fast inference model. Our proposed approach achieves approximately 95\% of the performance obtained through manual supervision by human annotators. The framework's performance was evaluated on a large sinkhole database, covering diverse sinkhole dateset-induced sinkholes using both aerial and high-resolution drone imagery. This paper presents the first self-supervised framework for sinkhole segmentation, demonstrating the robustness of foundational models (such as SAM and Depth Anything V2) when combined with prior topographic and geometry knowledge and an iterative self-learning pipeline. SinkSAM-Net has the potential to be trained effectively on extensive unlabeled RGB sinkholes datasets, achieving comparable performance to a supervised model. The code and interactive demo for SinkSAM-Net are available on the project page \href{https://osherr1996.github.io/SinkSAMNet}{at this URL}.

\end{abstract}


\end{frontmatter}


\section{Introduction}

Soil sinkholes have a profound impact on landform evolution and soil degradation \citep{yang2025mitigating}. Many studies indicate that sinkholes can lead to irreversible damage to built areas, cause roads collapse, destroy farmlands, and induce dam piping  \citep{SinkholeHazards}. Monitoring soil sinkholes occurrence, and characteristics, is crucial for studying dynamics of sinkhole-prone areas and mitigating sinkhole-related hazards. 

Sinkholes manifest on the Earth's surface as depressions, which can be observed either with the naked eye or through remote sensing (RS) platforms \citep{kim2019evolution}. However, an automatic detection of soil sinkholes using RS imagery is still a challenging task that was tackled by few researchers so far \citep{RAHIMI2024107711}. The reasons for this research gap may be the fact that soil sinkholes are characterized by irregular shapes, suffer internal and external shadowing effect, can be masked by vegetation, and are usually spatially distributed over wide regions representing a typical imbalanced data problem  \citep{bernatek2018subsurface}.

To tackle these challenges, ancillary topographic information may be used to improve sinkholes segmentation in RS imagery. Data sources such as LiDAR (light detection and ranging), or photogrammetric Digital Elevation Models (DEMs) 
\citep{PARDOIGUZQUIZA2021480, ARAV2025235} were used for various topographic analyses. However, airborne LiDARs are rarely being used at high-resolution over wide regions, due to extremely high costs \citep{raj2020survey}. Photogrammetric DEMs can have low accuracy in representing terrain, as they often mistakenly capture vegetation or roof structures as part of the ground surface. They are also sensitive to lighting and shadows, with poor conditions potentially causing errors in elevation data. Compared with LiDAR-derived DEMs, photogrammetric models typically have limited vertical accuracy. Additionally, they struggle to capture steep terrain, as overlapping image pairs may not fully represent hillslope sharp angles. Stereo matching issues can further result in gaps or distortions in the final DEM \citep{UYSAL2015539}.

As a substitute to these two topographic data sources, recently developed monocular depth estimation (MDE), using deep learning techniques, based on a single RGB image,  became central in the field of computer vision \citep{zhao2020monocular}. MDE was proven useful in 3D reconstruction and target instance segmentation. However, in RS studies, MDE is still very rarely used \citep{9372395}, particularly in geomorphological studies. This is unfortunate because MDE does not require multiple views as references, leading to improvements in both process time and efficiency \citep{MING202114}.

Automated sinkhole extraction was based on various geomorphometric variables, such as hillslope gradient, curvature, and topographic position \citep{ZHU2020125049}. The "fill sinks" method was successfully applied in various studies \citep{PARDOIGUZQUIZA2021480, HOFIERKA2018265}, but the features extracted by this approach often suffer from overestimation and low precision in delineating sinkhole boundaries. Consequently, depressions identified using topographic data frequently require refinement, usually through post-processing and visual inspection of RGB imagery by humans \citep{brazil}.

Besides geometric deterministic topographic computation, topographic information was used to enhance segmentation models, such as Convolutional Neural Networks (CNNs) \citep{article99}. Various encoder–decoder architectures were applied to sinkhole segmentation tasks \citep{8113128}, often using combinations of Digital Elevation Models (DEMs) and RGB imagery as inputs. 

Recently, foundational zero-shot models have gained traction in computer vision tasks \citep{Luddecke_2022_CVPR}, offering enhanced generalization by learning high-level concepts and relationships, rather than being restricted to predefined class labels learned during training \citep{liu2024groundingdinomarryingdino}. The ViT-based Segment Anything Model (SAM) enables image segmentation with minimal human intervention. SAM only requires a bounding box or a point as a prompt to initiate segmentation \citep{kirillov2023segment}. 
Nonetheless, sinkholes share common geometric features that can serve as strong priors for generating pseudo-labels using SAM to reduce human intervention during training \citep{DAROCHANUNESDECASTRO2024109212}. To date, all published studies on sinkhole segmentation rely on large volumes of fully supervised, human-annotated images for training, which is both costly and time-consuming and thus limits scalability when dealing with large RS datasets \citep{Jones2024}.

While SAM shows promise in serving as a zero-shot pseudo-label generator through interaction with sinkhole geometry as prompts, it still faces notable challenges in RS applications \citep{xiao2024foundationmodelsremotesensing}. Moreover, SAM incurs significant computational overhead due to its parameter-heavy architecture and still relies on prompt inputs to achieve reasonable accuracy, even after fine-tuning \citep{Ren_2024_WACV}. Consequently, there is a need for a more advanced, lightweight, and automated approach that leverages zero-shot SAM capabilities for self-supervised learning, enabling accurate sinkholes segmentation.

In response to these challenges, this paper proposes a knowledge-driven, self-supervised, general framework for sinkhole segmentation, under complex and volatile conditions, named the SinkSAM-Net framework. We aim to improve sinkhole segmentation through a novel, fully automatic, self-supervised, self-correcting, and computationally efficient lightweight architecture. Three operative objectives were set to achieve this aim: (1) We automatically refine DEM-derived, imperfect closed-depression boundaries using prompt-based segmentation of drone RGB imagery, through an innovative iterative coordinate-wise bounding box jittering label generator based on Monte Carlo simulations with zero-shot SAM 2.1; (2) We simplify the process to rely solely on RGB imagery by integrating deep learning-based Depth Anything V2 (DAV2) elevation estimation; Finally (3) we transfer knowledge from heavy rich in parameters SAM into a lightweight end-to-end learning and prompts free semantic segmentation architecture, incorporating an EfficientNetV2-based U-Net with a low parameter count and minimal computational requirements, to enhance the framework practical applicability. A comparative study against human-annotated, supervised datasets is conducted to further demonstrate the effectiveness of the proposed framework. Ultimately, we envision the SinkSAM-Net framework as a stepping stone toward the development of a foundation model for large-scale and generalizable sinkhole segmentation. 

The main contributions of this paper to the field of sinkhole segmentation are as follows: (1) SinkSAM-Net achieved pixel-level refinement of traditional DEM-based closed depression computation by incorporating RGB data with innovative coordinate-wise bounding box jittering (CWBJ) and shape-aware filtering to generate high-quality pseudo-labels;
(2) The use of automatic prompts derived from monocular depth estimation, based solely on a single RGB image, transcended limitations and biases of photogrammetric DEMs. This made sinkhole mapping feasible without relying on expensive LiDAR data. DAV2 demonstrated its effectiveness as an AI-based substitute for 3D elevation data in the "fill sinks" process, providing useful prompts for SAM; (3) The iterative self-training process with lightweight EffV2-UNet model further enhanced performance, ultimately resulting in a fast, robust, and generalizable sinkhole segmentation model.

\begin{table*}[ht!]
    \centering
    \caption{Comparison of Deep Learning Models for Sinkhole Detection Across Studies}
    \label{tab:sinkhole-segmentation-comparison}
    \footnotesize
    \begin{tabular}{p{2.7cm} >{\centering\arraybackslash}p{1.5cm} >{\centering\arraybackslash}p{4.3cm}
    >{\centering\arraybackslash}p{2.3cm} >{\centering\arraybackslash}p{0.3cm} >{\centering\arraybackslash}p{3cm} 
    >{\centering\arraybackslash}p{1.8cm}}
    \toprule
    \textbf{Study} & \textbf{Method} & \textbf{Input Data} & \textbf{Supervision} & \textbf{Sinkholes} & \textbf{Main Metrics} & \textbf{Origin} \\ 
    \midrule
    \citeauthor{rafique2022automatic}, \citeyear{rafique2022automatic} & U-Net & LiDAR DEM + Aerial Imagery & Supervised & 2,177 & IoU: 0.4538 & Karst \\
    \citeauthor{YAVARIABDI2023966}, \citeyear{YAVARIABDI2023966} & SinkholeNet & UAV RGB + Slope Maps & Weakly-Supervised & 62 & F1: 0.9708 & Karst \\
    \citeauthor{alrabayah2024deep}, \citeyear{alrabayah2024deep} & MC-U-Net & Satellite \& UAV RGB + DSM & Supervised & 1,288 & F1: 0.9123 & Salt dissolution \\
    \citeauthor{kariminejad2024evaluation}, \citeyear{kariminejad2024evaluation} & Res-U-Net & UAV RGB & Supervised & 203 & F1: 0.88 & Soil piping \\
    \citeauthor{jiang2024detection}, \citeyear{jiang2024detection} & PointNet++ & UAV RGB + LiDAR DEM & Supervised & 249 & IoU: 0.662 & Soil piping \\
    \citeauthor{10411919}, \citeyear{10411919} & AMFENet & LiDAR DEM & Supervised & 1,610 & IoU: 0.740 & Karst \\
    \citeauthor{cocskuner2025sinkhole}, \citeyear{cocskuner2025sinkhole} & YOLOv8 & Hillshade (DEM) & Supervised & 174 & F1: 0.870 & Karst \\
    \bottomrule
    \end{tabular}
    
    \vspace{0.5em}
    \begin{minipage}{0.95\textwidth}
    \footnotesize
    \textit{Note:} Res-U-Net = U-Net with ResNet encoder; MC-U-Net = Multi-Class U-Net.
    \end{minipage}
\end{table*}

\section{Related Work}

\subsection{Closed Depressions Computation for Sinkhole delineation}
Topographic depressions, also known as closed depressions, can be extracted by subtracting the elevation values of a sinks-free Digital Elevation Model (DEM) from those of the original DEM \citep{ZUMPANO2019213}. The sink-free DEM is typically generated using the “fill sinks” algorithm, as described by Planchon and Darboux (2001) \citep{PLANCHON2002159}, Wang and Liu (2006) \citep{wang2006efficient}, and others \citep{yong2009another}. A sink is defined as a surface cell surrounded by higher terrain, lacking a drainage path. The pour point is the lowest boundary cell where overflow would occur.

The Planchon and Darboux algorithm was implemented in ArcPy library and was applied in many studies, for different real world conditions and sinkholes origins \citep{PARDOIGUZQUIZA2021480,HOFIERKA2018265,article_photo}. For example, it was applied to detect karst sinkholes in Kentucky using LiDAR DEMs with 1 m resolution, achieving a detection rate of 97\% \citep{zhu2014improved}. "Fill sinks” was also applied to small soil sinkholes induced by soil piping, in conjunction with derivatives of airborne LiDARs. Success rate was satisfactory on grasslands/pastures (76\% for individual forms and 80\% for piping systems). \citep{BERNATEKJAKIEL2021107591}

Automatic sinkhole extraction methods from DEMs have two main limitations, both result in from missing RGB information: (1) DEM-based methods tend to delineate non-sinkhole features that have a similar geometric representation in DEMs \citep{HOFIERKA2018265}; and (2) Sinkholes delineated from DEMs often have inaccurate boundaries \citep{chen2015semi}, hampered by vegetation, mainly shrubs, within the sinkholes. Consequently, DEMs often do not sufficiently represent the actual sinkhole structure \citep{LINDSAY20061192}. To improve DEM-based methods, researchers often filter out such noisy features in post-processing with visual inspection of RGB imagery by human annotators \citep{brazil}.

\subsection{Deep Learning Sinkhole Segmentation}

Various encoder-decoder deep learning networks have been applied to sinkhole segmentation. Input data included RGB images, DEMs and their derivatives, or combinations of the two sources \citep{alrabayah2024deep}. The U-Net architecture, in particular, allows for effective multiscale feature capture. For example, the implementation of a classic U-Net architecture has resulted in an IoU of 0.4538 and a Precision of 0.6629 for karst sinkhole segmentation using LiDAR DEM and aerial imagery with a spatial resolution of 1.5 m, across a large dataset of 2,177 sinkholes \citep{rafique2022automatic}. It was also found that elevation gradient images, as inputs, outperformed RGB imagery, as the latter provided weak cues for segmenting sinkholes.

When applied to salt-induced, well-defined sinkholes in drylands, an improved multi-class U-Net using RGB-only UAV imagery performed promisingly, achieving a F1 score of 0.9123 \citep{alrabayah2024deep}. However, for small soil sinkholes in semiarid regions, automatic segmentation of sinkholes from UAV LiDAR data achieved relatively low scores, with an IoU of 0.662 \citep{jiang2024detection}.

Beyond the U-Net architecture \ref{kariminejad2024evaluation}, more advanced modifications were introduced to increase sinkhole segmentation and detection accuracy. For example, AMFENet incorporates Adaptive Multiscale Feature Fusion Enhancement into the U-Net framework by using the Swin Transformer as its encoder. This architecture was trained and tested on LiDAR DEM data in Kentucky karst regions to capture contextual global information, achieving an IoU of 0.740 on an extensive test dataset of 1,610 sinkholes \citep{10411919}. Another dual-stream network, based on inputs of RGB and hillslope gradient data using improved ShuffleNet architecture and gradient-weighted class activation mapping (Grad-CAM) with weak supervision, was trained using human-labeled patches indicating whether a sinkhole was present or not, resulting in a binary classification task. Although the method is novel and achieved a high F1-score of 0.9708, it was tested on a small dataset of only 62 sinkholes instances and focused on patch-level classification with localization rather than  segmentation \citep{YAVARIABDI2023966}. A more recent approach employed the state-of-the-art YOLOv8 for karst sinkhole detection, yielding an F1-score of 0.870 on a dataset of 174 sinkholes\citep{cocskuner2025sinkhole}.

Still, the predominant learning paradigm within the domain of sinkhole segmentation and detection remains fully supervised learning based on human-annotated data. SAM has not yet been applied to sinkhole segmentation tasks, and all existing datasets rely on photogrammetric or LiDAR-derived information (Table~\ref{tab:sinkhole-segmentation-comparison}).

\subsection{Monocular Depth Estimation}
Deep Learning-Based Monocular Depth Estimation (MDE) model allows to quantify depth from a single RGB image using training sets \citep{zhao2020monocular}. Most MDEs use an encoder–decoder architecture to minimize loss in training \citep{8359371}. Depth estimation relies on cues such as: shading, occlusion, perspective, texture variations, and object scaling, to differentiate and understand the scene \citep{godard2017unsupervised}. Monocular depth estimation was used for 3D reconstruction purposes, augmented reality, autonomous driving, and robotics \citep{MING202114}.

Due to the limited payload capacity of most drones, they are typically outfitted with only a single camera. This prevents a widespread use of depth perception methods based on LiDAR or photogrammetric DEMs \citep{10.1007/978-3-319-10605-2_7}. As an alternative, with the success of CNNs, MDEs became increasingly reliable \citep{DBLP:journals/corr/LiuSL14}. MDEs have gained increasing attention in remote sensing (RS), although most existing studies have focused on urban environments or forests, primarily aiming at scene reconstruction involving buildings or tree canopies \citep{9372395, isprs-annals-V-2-2020-451-2020, 10802125}. By contrast, MDEs are still rarely used \citep{9372395}, in studies of open environments and particularly of geomorphological processes in applications such as hillslope erosion processes, stream networks and sinkholes extractions.

Pre-trained Foundation Models have gained popularity in MDE tasks, being trained using large datasets and enabling zero-shot or fine tuning predictions \citep{birkl2023midasv31model}. In particular, DAV2 \citep{yang2024depthv2} introduces key advantages: (1) It overcomes limitations of image annotation by involving synthetic annotated data during training with fine detailed annotation; (2) It uses a teacher-student architecture to obtain a lightweight and accurate model. The model is trained on synthetic images. Then the teacher mode generates pseudo masks on unannotated real images. Finally, the smaller student model is trained on these pseudo-annotated images; (3) Its annotation pipeline incorporates  SAM for image sampling, utilizing a combination of real- and pseudo-labeled depth maps, from 62 million images \citep{yang2024depthanythingunleashingpower}. 

DAV2 is benchmarked on standard datasets e.g., KITTI, Sintel, and more, outperforming previous Depth Anything version and the MiDas V3.1 from intel labs research \citep{birkl2023midasv31model}. A key highlight is DAV2's superior performance, when evaluated on the DA-2K dataset, which includes eight different image categories, 9\% of which are aerial imagery and fine-detail images \citep{yang2024depthv2}. This indicates potential for terrain elevation estimation also from drone imagery. DAV2 was applied in a few studies of the open environment \citep{zhang2024towards}, for example, for agricultural canopy height measurements \citep{cambrin2024depth, 10978076},and in urban environment  providing a high performance solution, surpassing the current state-of-the-art.

\begin{figure*}[ht!]
    \centering
    \includegraphics[width=1\linewidth]{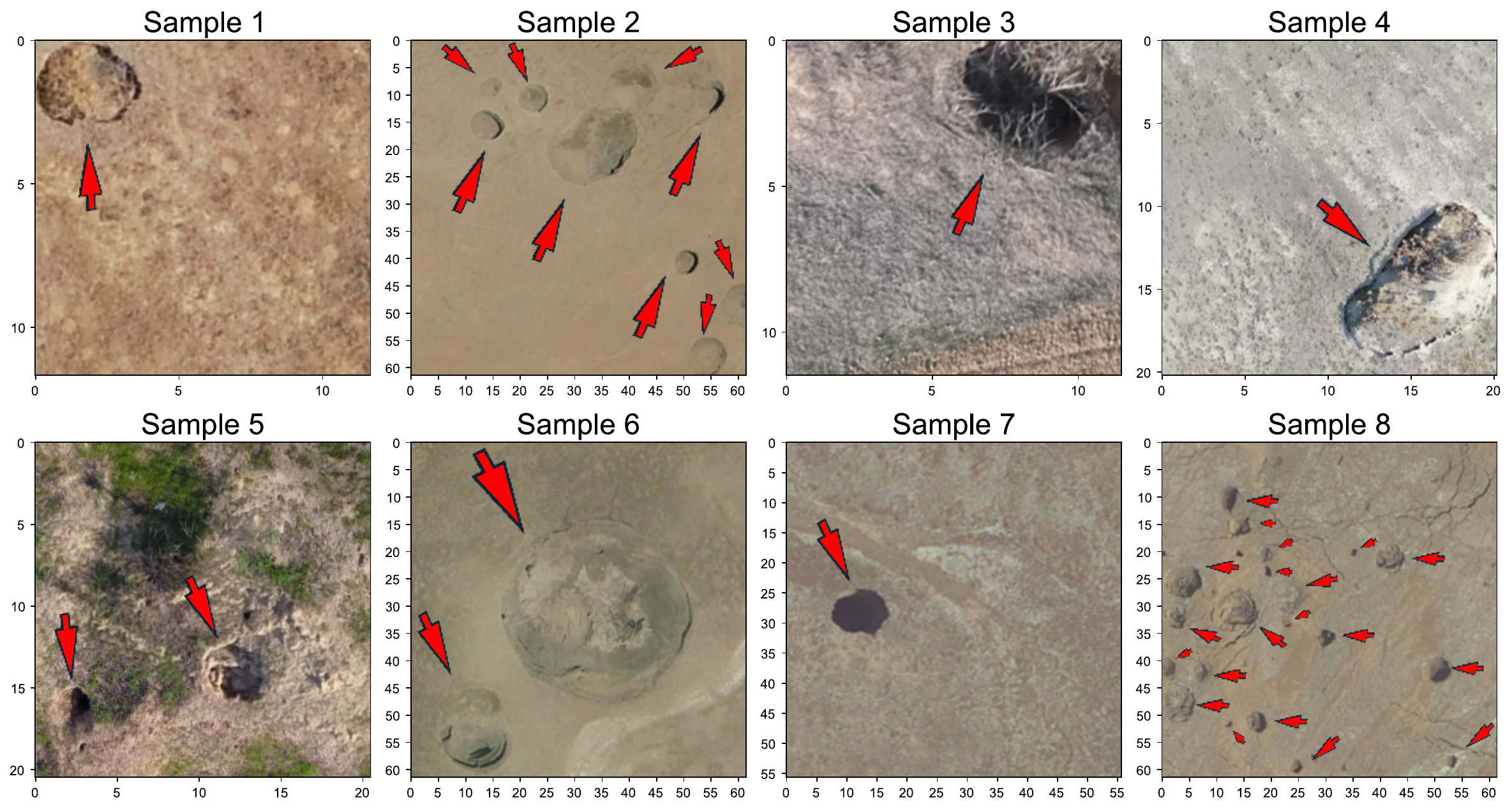}
    \caption{Sample images from the sinkhole dataset used here: This collection includes 8 samples representing different sinkholes captured using RS platforms. UAV-based sinkhole images with a spatial resolution of 3–5 cm per pixel (samples 1, 3, 4, 5), while aerial imagery varies between 13–17 cm per pixel. The images were captured in different climatic regions: semi-arid during the wet season (sample 5), hyper-arid (samples 2, 7, 8) and semi-arid during the dry season (samples 1, 3, 4). The sinkholes also represent various origins, including small soil sinkholes (samples 1, 3, 5), large karst sinkholes (sample 4), and large salt-induced sinkholes (samples 2, 6, 7, 8).}
    \label{dataset}
\end{figure*}

\subsection{Segment Anything Model}

SAM is an encoder–decoder image-segmentation model. Its encoder is a pre-trained Vision Transformer (ViT) whose weights and biases were learned from the 1 Billion Mask (SA-1B) dataset \citep{kirillov2023segment}. Unlike traditional encoder–decoder architectures, SAM incorporates a prompt encoder that maps user input such as a point or a bounding box into an embedding vector, which is processed together with the image. The model offers three principal advantages: (1) promptable segmentation, which yields high-quality object masks from simple prompts; (2) training on a large-scale dataset of 11 million images and 1.1 billion masks that were automatically generated by the model itself; and (3) strong zero-shot performance across a wide range of segmentation tasks.

SAM was applied to image processing tasks, including medical imaging \citep{10386032}, shadow segmentation \citep{zhang2023samhelpsshadowwhensegmentmodelmeet}, electron microscopy segmentation \citep{Archit2023.08.21.554208}, crack segmentation \citep{ahmadi2023application}, and road pit detection. In RS, SAM was also used for geological features analysis on planetary bodies \citep{GIANNAKIS2024115797}, glacier segmentation \citep{shankar2023semantic}, road mapping \citep{10613866}, building segmentation \citep{10636322}, and solar panel segmentation \citep{rafaeli2024promptbasedsegmentationmultipleresolutions}.

SAM models are trained and tested mostly on ground photographs and videos \citep{ravi2024sam}. Recently, the improved SAM 2.1 version was released. Still, due to the unique properties of RS images, SAMs faces challenges when dealing with small objects or imperfect prompts. SAM may overestimate object size by including shadows in the segmented regions \citep{sultan2024geosamfinetuningsamsparse}. Thus, in RS images, without re-training, fine-tuning, or prompt manipulation, SAMs may not achieve the same level of accuracy as algorithms specifically designed for RS segmentation tasks \citep{WEI2024446, WANG2024707}. Among the proposed solutions, trainable prompt generation modules, such as RSPrompter, have achieved notable improvement in SAM's segmentation performance on remote sensing applications \citep{chen2023rsprompterlearningpromptremote}. However, these methods require supervised training.

To address the unique challenges of remote sensing while preserving the zero-shot setting suitable for self-supervised learning, a prompt augmentation technique can be adopted to generate a series of input prompts for SAM, aimed at uncertainty estimation \cite{zhou2024medsamuuncertaintyguidedautomultiprompt}. An uncertainty-based rectification module was used to further leverage the distribution of estimated uncertainty to improve segmentation performance \citep{zhao2024iterative, 10.1007/978-3-031-47425-5_33}. This method, without supplementary training or fine-tuning, improved segmentation performance, increasing the dice similarity coefficient by up to 10.7\% and 13.8\% across medical images of different organs \citep{DBLP:journals/corr/abs-2311-10529}. This approach also has potential in RS applications, especially when prompts are inaccurate or incomplete. By applying small jittering, SAM can provide segmentation even when the bounding box is not perfectly localized around the entire object. Moreover, when no object is present, SAM’s uncertain predictions may allow such prompts to be ignored.

The zero-shot capabilities of SAM also make it a strong candidate for pseudo-label generation in self-supervised learning frameworks \citep{jiang2023segmentgoodpseudolabelgenerator, yang2024samguidemasteringpseudolabel}. These frameworks are primarily guided by prior domain knowledge \citep{10.1007/978-981-99-8141-0_11}, often relying on pixel values and classical image processing techniques. For example, dark pixel values representing soil cracks were combined with SAM outputs to generate high-quality segmentation labels \citep{xu2024soil}.

While foundation models (e.g. ViT-based SAM) demonstrate strong generalization capabilities, they often suffer from high computational costs for large-scale or high-resolution imagery. Even though Selective State Space Models (SSMs), as used in models like DynamicVis, significantly improve efficiency \citep{chen2025dynamicvisefficientgeneralvisual}, they still require considerable computational resources and complex training pipelines \citep{huang2024stochasticlayerwiseshufflegood}.

Conversely, lightweight CNNs provide faster training, fewer parameters, and lower inference latencies when used with moderate to large-size datasets that have clear spatial structures \citep{tan2020efficientnetrethinkingmodelscaling}. Given these advantages, we adopted a CNN-based architecture for the target model, as it aligns well with our task-specific requirements \citep{he2024mobilemambalightweightmultireceptivevisual}.

\section{Materials and Methodology}

\begin{figure*}[ht!]
    \centering
    \includegraphics[width=1\linewidth]{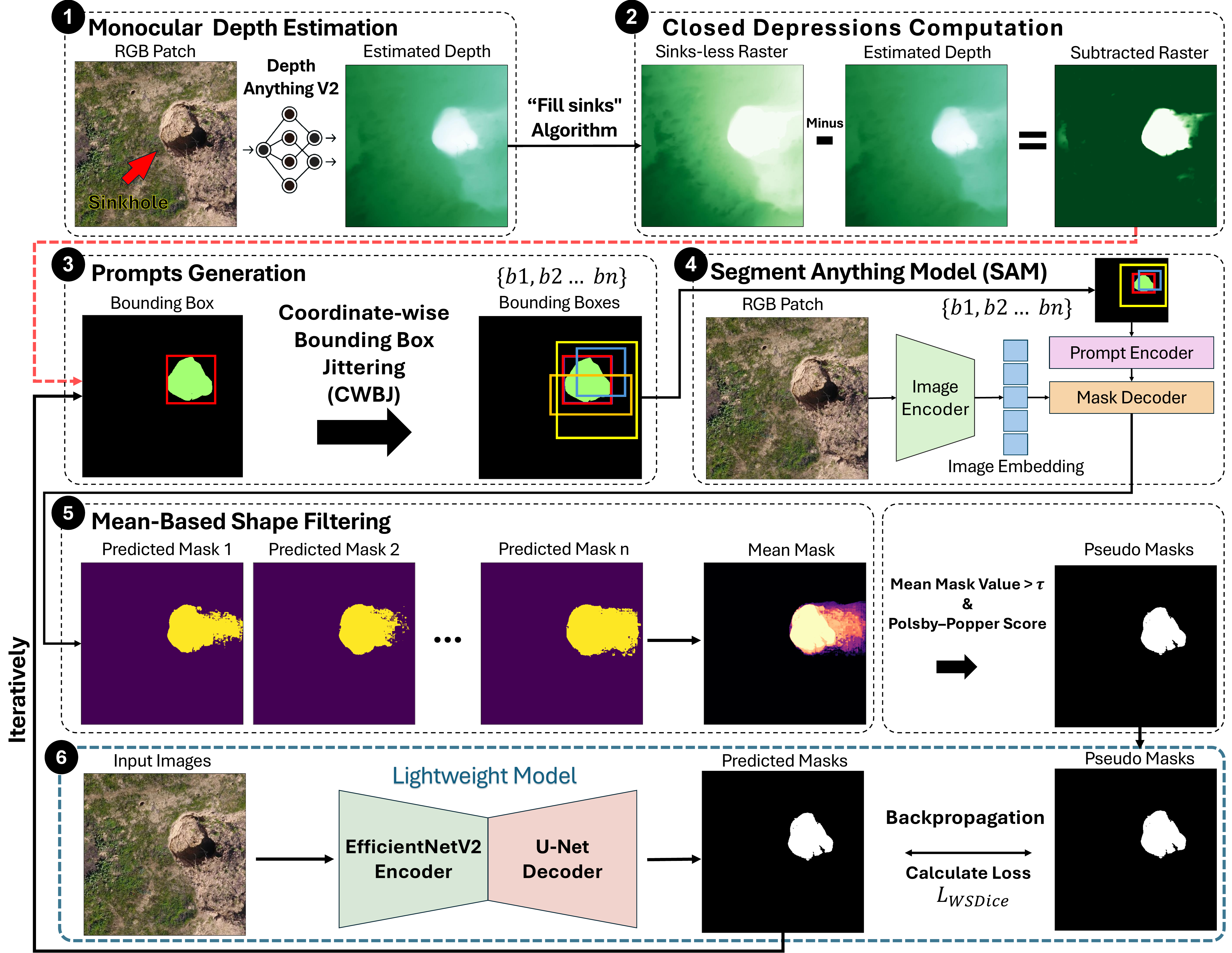}
    \caption{SinkSAM-Net framework: \textit{Stage 1}: Depth estimation from an RGB image using DAV2; \textit{Stage 2}: Application of the "fill sinks" technique and subtraction of the estimated depth from a sink-free raster to delineate closed depressions; \textit{Stage 3}: Prompt generation employs coordinate-wise bounding-box jittering (CWBJ), which is applied to each bounding box to create a series of boxes for Monte Carlo segmentation simulations; \textit{Stage 4}: SAM model uses an image encoder and mask decoder to generate a series of sinkhole masks corresponding to the jittered bounding boxes; \textit{Stage 5}: For mean-based shape filtering, the patch masks are first averaged to produce a mean mask; this mask is then filtered according to confidence levels and geometric properties evaluated with the Polsby–Popper circularity index; \textit{Stage 6}: Final training of the lightweight end-to-end EffV2-UNet target model using the filtered pseudo-labels.(The red-dashed line between Stage 2 and Stage 3 indicates that The closed depressions (CDs) were first fed into the SAM model, and later, SAM was iteratively guided by the outputs of the segmentation model. }
    \label{framework}
\end{figure*}

\subsection{Study Area and Dataset}

The framework suggested here was applied to data from three different regions, combined into a single sinkhole dataset, sourced from multiple sinkhole collections spanning different years, seasons, RS platforms, origins, and geographic areas. The data derived from RS imagery capturing sinkholes across three regions:
(1) The northwestern Negev Desert, east of the Mediterranean Sea coastline of Israel, characterized by a semi-arid climate \citep{IsraelMeteorologicalService} and sandy loam/silt loam soil textures \citep{STAVI201688}, soil piping sites including 500 sinkhole instances;
(2) Salt-induced sinkholes near the Dead Sea in a hyper-arid climate, also in Israel, comprising 2,038 sinkhole from four years;
(3) Karst sinkholes located in Konya, Turkey, with 62 sinkhole instances \citep{YAVARIABDI2023966}. The annotation was performed using GIS tools by an expert with extensive experience in sinkhole annotation. 

Sinkholes in the dataset vary in morphology, lighting, shadowing, size, vegetation, and imaging resolution and conditions across regions and acquisition dates (Fig.~\ref{dataset}). RGB imagery was collected using drones over soil piping sinkhole catchments and the Konya region, with spatial resolutions of 3–5 cm per pixel \citep{YAVARIABDI2023966}. In contrast, imagery from the Dead Sea region was acquired using aerial platforms at a coarser resolution of 13–15 cm per pixel.

In total, our dataset contains 2,600 sinkhole samples captured from multiple regions over the period 2017–2023. All orthomosaics were divided into patches of $512^2,\text{px}^2$ and randomly split into training/validation ($N = 1800$, ~70\%) and test ($N = 800$, ~30\%) sets. To avoid data leakage in areas with multiple acquisitions across different years, the random split was performed by geographic location, ensuring that patches from the same location were assigned to only one set (training/validation or test) while still preserving the overall dataset split ratio.


In addition to RGB data, soil piping sites were captured with 60\% overlap to produce a DEM, using photogrammetry Agisoft Metashape with a pixel length of 0.025 m. 

\subsection{SinkSAM-Net Framework Overview}

The SinkSAM-Net framework for sinkhole segmentation developed in the current work is illustrated in Figure \ref{framework}. The framework consists of seven stages:
\textit{Stage 1}: Depth estimation using DAV2 from a single RGB image.
\textit{Stage 2}: Delineation of closed depressions, done by: (1) executing the "fill sinks" algorithm; and (2) subtracting the original depth raster from the "filled sinks" raster, using depth data from Stage 1.
\textit{Stage 3}: Prompts are generated by first applying a threshold to remove small sinks; the remaining depressions from Stage 2 are then converted into a series of bounding boxes using coordinate-wise bounding-box jittering for SAM implementation in Stage 4.
\textit{Stage 4}: SAM utilizes an image encoder and mask decoder to segment sinkholes.
\textit{Stage 5}: All segmentation masks are saved and used to compute a mean mask, which is then thresholded using a confidence level and geometric constraints.
\textit{Stage 6}: A lightweight segmentation model is trained using unlabeled images and the pseudo-labels generated in Stage 6. This trained model iteratively enhances the pseudo-labels and retrains the target model.

\textbf{Stage One – Monocular Depth Estimation:}
Orthomosaic patches were processed using the pre-trained DAV2 model. A large Vision Transformer (ViT-L) encoder was used to achieve optimal accuracy in depth estimation. The patches preserved the same resolution and geographic data as the original RGB images to ensure accurate alignment. Based on simulations, we found the patch size of $512^2, \text{px}^2$ as optimal for DAV2 estimation. 

\textbf{Stage Two – Closed Depressions Computation:}
The ArcPy "Fill" tool \citep{desktop2011release} was applied to fill all closed depressions in the DAV2 surface, up to their spill elevation (i.e., the elevation at which water would ideally flow out of the depression). This created a depression-less depth map. The original DAV2 depth data was then subtracted from the depression-less raster to generate a difference raster representing depression location and depth.

\textbf{Stage Three – Prompts Generation:}
The subtracted layer was filtered using two threshold values to remove small depressions. The threshold values were: depth \textless 2. Closed depressions in each patch were used to generate primary bounding boxes. On each bounding box, we applied coordinate-wise bounding box jittering (CWBJ), randomly moving each corner independently by 10 px. These jittered boxes were used as SAM inputs along with the corresponding RGB data.

\textbf{Stage Four – Segment Anything Model (SAM):}
RGB patches were processed by SAM using the corresponding series of prompt boxes extracted from the initial depression layer (Stage 2). Masks were generated for each bounding box separately, and then all predictions were combined. If more than a single object appeared in a single image, predictions were merged by selecting pixels with the highest foreground probability.

\textbf{Stage Five – Mean-Based Shape Filtering:}
Patch masks were processed to generate a mean mask. Consistently predicted pixels yield high values, close to 1, while unstable or inconsistently predicted regions vary between 0 and 1 and yield moderate mean values. By thresholding the mean mask, we ensure that repeatedly predicted regions likely correspond to well-defined sinkholes, while unstable predictions, possibly corresponding to non-object regions, are filtered out. After thresholding, boundaries are refined. We then apply the Polsby–Popper score \citep{polsby1991third} to filter out non-circular shapes see algorithm \ref{alg:mc_segmentation_cwbj}. Based on prior knowledge, we know that sinkholes often exhibit Polsby–Popper scores greater than 0.5 \citep{RAFAELI2023107511}.

\begin{equation}
\text{PP}(R) = \frac{4\pi \cdot A_R}{P_R^2}
\label{eq:pp}
\end{equation}

\noindent
where \( \text{PP}(R) \) denotes the Polsby--Popper compactness score for a segmented region \( R \), representing a single sinkhole. Here, \( A_R \) is the area of the region, and \( P_R \) is its perimeter. The score ranges from 0 to 1, with higher values indicating more compact (circular-like) shapes and lower values reflecting irregular or elongated geometries.

\textbf{Stage six – Lightweight Segmentation Model:}
At this stage, we begin training our end-to-end lightweight target model, fed during training with RGB images and the pseudo-labels filtered in Stage 5. After the model is trained, we predict again on the training dataset, pass the labels through SAM using the CWBJ mechanism, and retrain the target model. This process is repeated twice.

\begin{algorithm}[ht!]
\caption{Monte Carlo Segmentation with Shape-Aware Filtering followed CWBJ}
\label{alg:mc_segmentation_cwbj}
\begin{algorithmic}[1]
\Statex \textbf{Inputs:} Input image $I$, bounding boxes $B$, SAM $\mathcal{M}$, number of samples $N = 50$, threshold $\tau = 0.5$, Polsby–Popper threshold $PP_{\text{min}} = 0.3$
\Statex \textbf{Output:} Mean mask $\mu$, filtered mask $M_f$

\State Initialize an empty list of masks: $\mathcal{S} \gets [\,]$
\For{$n = 1$ to $N$}
    \State Initialize $M_n$ as an empty mask
    \For{each bounding box $b \in B$}
        \State $b' \gets \text{JitterBox}(b)$ \Comment{Apply small random shifts}
        \State $m \gets \mathcal{M}(I, b')$ \Comment{Run model on jittered box}
        \State $M_n \gets \max(M_n, m)$
    \EndFor
    \State Append $M_n$ to $\mathcal{S}$
\EndFor

\State Compute mean mask: $\mu \gets \frac{1}{N} \sum_{i=1}^{N} \mathcal{S}_i$
\State Binarize: $\hat{M} \gets \text{Binarize}(\mu, \tau)$
\State Label connected regions: $\mathcal{R} \gets \text{LabelComponents}(\hat{M})$
\State Initialize $M_f \gets 0$

\For{each region $r \in \mathcal{R}$}
    \If{$\text{Area}(r) < 50$}
        \State \textbf{continue}
    \EndIf
    \State $PP \gets \text{PolsbyPopper}(r)$ \Comment{Eq.~\eqref{eq:pp}}
    \If{$PP > PP_{\text{min}}$}
        \State Add $r$ to $M_f$
    \EndIf
\EndFor

\Return $M_f$
\end{algorithmic}
\end{algorithm}

\begin{table}[ht!]
\centering
\caption{Configuration of the proposed EffV2-UNet.}
\label{tab:effnetv2unet}
\footnotesize
\begin{tabular}{p{1.9cm} p{2cm} p{3.7cm}}
\toprule
\textbf{Stage} & \textbf{Feature Size} & \textbf{Operation} \\
\midrule
Input           & 512 × 512 × 3     & – \\
Stem            & 256 × 256 × 24    & Conv + BN + ReLU \\
Encoder Stage 1 & 128 × 128 × 32    & Fused-MBConv \\
Encoder Stage 2 & 64 × 64 × 48      & MBConv + SE + Add \\
Encoder Stage 3 & 32 × 32 × 64      & MBConv + SE + Add \\
Encoder Stage 4 & 16 × 16 × 160     & MBConv + SE + Add \\
Decoder Stage 4 & 32 × 32 × 128     & Upsample + Concat + Conv ×2 \\
Decoder Stage 3 & 64 × 64 × 64      & Upsample + Concat + Conv ×2 \\
Decoder Stage 2 & 128 × 128 × 32    & Upsample + Concat + Conv ×2 \\
Decoder Stage 1 & 256 × 256 × 16    & Upsample + Concat + Conv ×2 \\
Output          & 512 × 512 × 1     & Upsample + 1×1 Conv (sigmoid) \\
\bottomrule
\end{tabular}

\vspace{1mm}
\begin{minipage}{\linewidth}
\footnotesize \textit{Note:} Feature map unit is pixel (px). BN stands for Batch Normalization. SE denotes Squeeze-and-Excitation module.
\end{minipage}
\end{table}

The proposed EffV2-UNet architecture is a lightweight encoder–decoder network that captures semantic representations through stacked convolutions and pooling operations, followed by upsampling to restore the original image size. Our U-shaped network integrates EfficientNetV2 blocks for efficient feature extraction and accurate segmentation, leveraging Neural Architecture Search (NAS) to optimally adjust the depth and width for maximum accuracy and efficiency \cite{tan2021efficientnetv2smallermodelsfaster}. In our proposed model, the encoder consists of a stem block followed by four stages: one Fused-MBConv and three MBConv blocks with Squeeze-and-Excitation (SE) and residual connections, all derived from the EfficientNetV2B0 network. The kernel size was manually set to 3×3, while the depth (number of layers per stage) and width (number of channels per layer) were automatically searched to balance accuracy and efficiency. The UNet decoder mirrors the encoder with four upsampling stages, each using 2×2 upsampling, concatenation with corresponding encoder features, and two 3×3 convolutional layers. The final output is generated through a 1×1 convolution followed by a sigmoid activation function (see configuration in Table~\ref{tab:effnetv2unet}). This design enables robust multi-scale representation while remaining computationally efficient.

During the intermediate stage, we tested the base U-Net \citep{DBLP:journals/corr/RonnebergerFB15} and DeepLabV3+, a fully Convolutional Neural Network (CNN) architecture for robust semantic segmentation tasks \citep{chen2018encoder}, both using a two lightweight ResNet18 and EfficientNetV2B0 backbones \citep{he2015deepresiduallearningimage}. Our proposed U-Net with the EfficientNetV2 configuration outperformed all other architectures and was chosen to serve as our target model. The intermediate results can be found in Table~\ref{tab:segmentation-iou-f1}.

\begin{table}[ht!]
    \centering
    \caption{F1 Score and Sinkhole IoU of UNet and DeepLab Variants Trained on Human-Labeled Data.}
    \label{tab:segmentation-iou-f1}
    \footnotesize
    \begin{tabular}{p{3cm} >{\centering\arraybackslash}p{2.2cm} >{\centering\arraybackslash}p{2.2cm}}
    \toprule
    \textbf{Model} & \textbf{Sinkhole IoU ↑} & \textbf{F1 Score ↑} \\
    \midrule
    UNet                  & 0.4490 & 0.5510 \\
    Res-UNet              & 0.5477 & 0.6485 \\
    EffV2-UNet            & \textbf{0.5508} & \textbf{0.6517} \\
    \midrule
    DeepLabV3+            & 0.4197 & 0.5340 \\
    DeepLabV3+ ResNet     & 0.4755 & 0.5766 \\
    DeepLabV3+ EffNetV2   & 0.5214 & 0.6221 \\
    \bottomrule
    \end{tabular}
\end{table}

\begin{figure*}[ht!]
    \centering
    \includegraphics[width=1\linewidth]{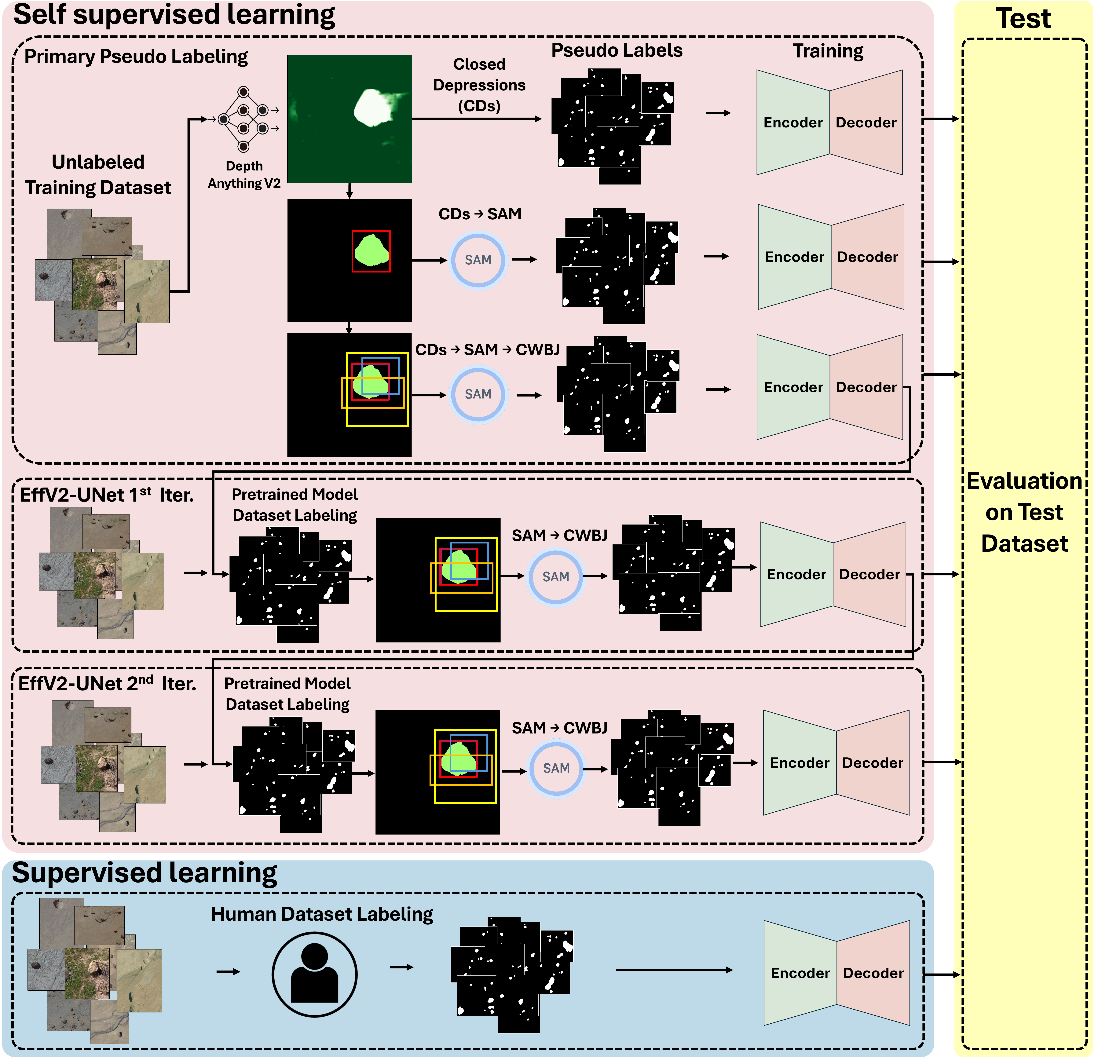}
    \caption{Experimental Setup: During the experimental setup, we trained the target model using different sets of pseudo-labeled data. First, we trained the EffV2-UNet model using closed depressions (CDs). Next, we passed the CDs through SAM to generate new masks and retrained the model. We then applied coordinate-wise bounding box jittering (CWBJ). Finally, we used the previously trained EffV2-UNet to relabel the dataset and repeated this self-training process twice. All trained models, were evaluated on the same test dataset. Lastly, we compared the full self-supervised pipeline with a conventional supervised learning approach using human-labeled data.}
    \label{new_exp}
\end{figure*}

\subsection{Implementation Details}
\textit{Training Setting:} SAM 2.1 large version was used to generate zero-shot segmentation predictions. All CNN-based models were implemented and trained using TensorFlow and Keras libraries \citep{tensorflow2015-whitepaper}. The implementation of our framework leveraged high-performance computing resources to ensure efficient training and inference of segmentation models. Specifically, we employed the Ultralytics v8.3.65 framework \cite{Jocher_Ultralytics_YOLO_2023} with PyTorch 2.3.1 for YOLO models, and Keras from TensorFlow for training U-Net and EfficientNet-based architectures. All experiments were conducted using Python 3.9.15 on an NVIDIA RTX A4000 GPU. 

Each model was trained for 35 epochs with a batch size of 8, using the Adam optimizer and a learning rate of \(10^{-3}\). For the human-labeled dataset, we found that Binary Cross-Entropy (BCE) loss yielded the best performance. For pseudo-labeled data, the best results were obtained using the Weighted Soft Dice (WSDice) loss \citep{wang2020improved}.

\textit{Loss Function:}  
A combination of the Binary Cross-Entropy Loss \( \mathcal{L}_{\text{BCE}} \) and Weighted Soft Dice Loss \( \mathcal{L}_{\text{WSDice}} \) was used for training the SAM mask decoder and CNN segmentation models. The WSDice loss is defined as:

\begin{equation}
\mathcal{L}_{\text{WSDice}} = 1 - \dfrac{2 \cdot \sum_i w_s y_i \hat{y}_i + 1}{\sum_i \left( w_s y_i + w_n (1 - y_i) \hat{y}_i \right) + 1}
\end{equation}

where \( y_i \) and \( \hat{y}_i \) denote the ground truth and predicted probability at pixel \( i \), respectively. \( w_s = 1.0 \) and \( w_n = 0.07 \) are the weights for the foreground (positive) and background (negative) classes, respectively. This weighting emphasizes foreground pixels (i.e., sinkholes), helping to counter class imbalance in segmentation tasks.

The Binary Cross-Entropy (BCE) loss is defined as:

\begin{equation}
\mathcal{L}_{\text{BCE}} = - \dfrac{1}{N} \sum_{i=1}^{N} \left[ y_i \log(\hat{y}_i) + (1 - y_i) \log(1 - \hat{y}_i) \right]
\end{equation}

where \( N \) is the total number of pixels in the image, and \( y_i \), \( \hat{y}_i \) represent the true and predicted values, respectively.

\textit{Evaluation Metrics:} To assess model performance, we employed the widely recognized evaluation metrics in Eq.~\eqref{eq:ba}--\eqref{eq:iou}:

\begin{equation}
\text{Balanced Accuracy} = \frac{1}{2} \left( \frac{TP}{TP + FN} + \frac{TN}{TN + FP} \right)
\label{eq:ba}
\end{equation}

\begin{equation}
\text{FPR} = \frac{FP}{FP + TN}
\label{eq:fpr}
\end{equation}

\begin{equation}
\text{FNR} = \frac{FN}{FN + TP}
\label{eq:fnr}
\end{equation}

\begin{equation}
\text{Precision} = \frac{TP}{TP + FP}
\label{eq:precision}
\end{equation}

\begin{equation}
\text{Recall} = \frac{TP}{TP + FN}
\label{eq:recall}
\end{equation}

\begin{equation}
\text{F1 Score} = 2 \times \frac{\text{Precision} \times \text{Recall}}{\text{Precision} + \text{Recall}}
\label{eq:f1_score}
\end{equation}

\begin{equation}
\text{IoU} = \frac{|A \cap B|}{|A \cup B|}
\label{eq:iou}
\end{equation}
where TP, TN, FP, and FN denote, respectively,  true positive, true negative, false positive, and false negative values, and \(A\) and \(B\) are predicted and ground truth segments, respectively. 

Balanced Accuracy accounts for class imbalance by averaging the recall of each class. FPR (False Positive Rate) and FNR (False Negative Rate) quantify model errors by describing the proportion of negative and positive samples misclassified, respectively. The F1-score and IoU (Intersection over Union) represent the semantic segmentation performance of the entire map after merging the individual patches back into the full image. A confusion matrix is used to evaluate the model's object detection performance by comparing predicted mask bounding boxes with ground truth bounding boxes at various IoU thresholds. It quantifies the detection rate at each IoU threshold, categorizing predictions into True Positives (TP), False Positives (FP), and False Negatives (FN) based on whether the IoU between predicted and ground truth masks meets or exceeds the threshold.

\section{Results}

\begin{figure*}[ht!]
   \centering
    \includegraphics[width=1\linewidth]{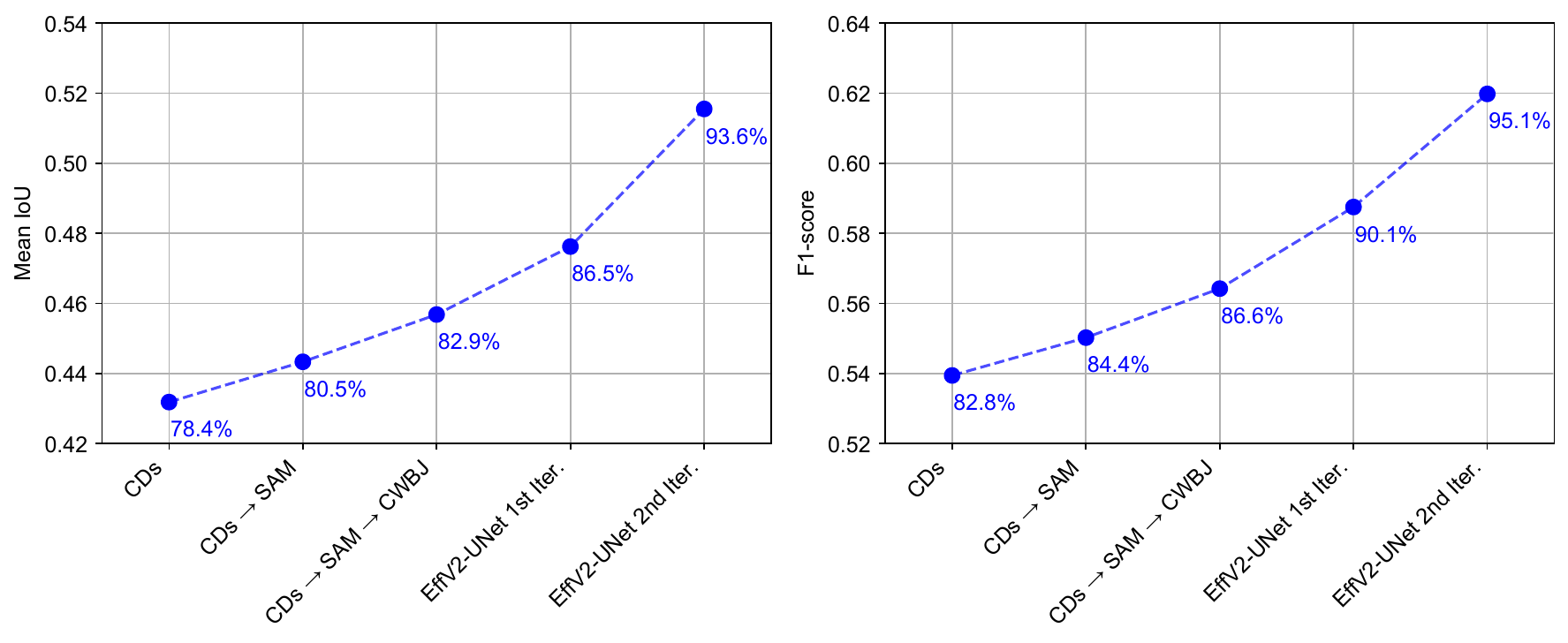}
    \caption{Different Pseudo-Label Origins vs. Model Performance:
    Each component of the SinkSAM-Net framework incrementally improves segmentation performance, ultimately reaching 95\% of the F1-score and 93\% of the IoU achieved by a fully supervised model trained on human-labeled data.}
    \label{maps}
\end{figure*}

\begin{figure*}[ht!]
   \centering
    \includegraphics[width=1\linewidth]{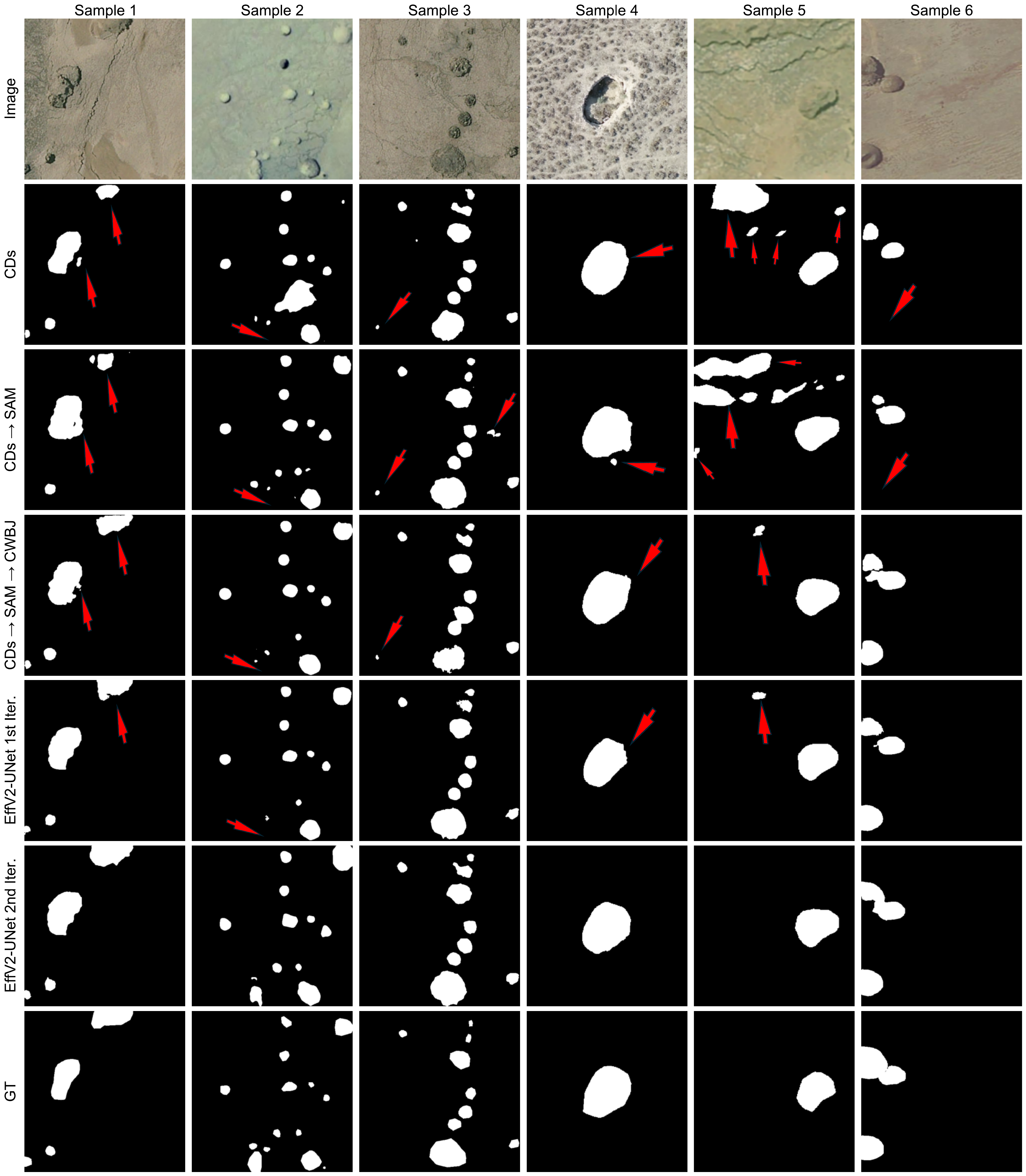}
    \caption{Visual Examples from the Evaluation of Trained Models with Different Pseudo-Labeling on Test Dataset:
    The red arrows highlight the main visual differences between the models. The best performance was achieved after the second iteration of EffV2-UNet training. The EffV2-UNet model segments sinkholes more completely, including those located at the edges of image patches. Additionally, it produces fewer false positives by reducing the number of incorrectly segmented background pixels.}
    \label{comparison_grid_models}
\end{figure*}

\begin{table*}[ht!]
    \centering
    \caption{Performance of the EffV2-UNet model trained on various pseudo-labels sources.}
    \label{tab:full-metrics-comparison}
    \footnotesize
    \begin{tabular}{p{4cm} >{\centering\arraybackslash}p{2cm} >{\centering\arraybackslash}p{1.5cm} >{\centering\arraybackslash}p{1.5cm} >{\centering\arraybackslash}p{1.5cm} >{\centering\arraybackslash}p{1.2cm} >{\centering\arraybackslash}p{1.2cm} >{\centering\arraybackslash}p{2cm}}
    \toprule
    \textbf{Training Dataset (Pseudo-labels)} & \textbf{Sinkhole IoU ↑} & \textbf{F1 Score ↑} & \textbf{Precision ↑} & \textbf{Recall ↑} & \textbf{FPR ↓} & \textbf{FNR ↓} & \textbf{Balanced Acc. ↑} \\ 
    \midrule
    CDs & 0.4318 & 0.5394 & \textbf{0.6468} & 0.5618 & 0.0110 & 0.4382 & 0.7754 \\
    CDs → SAM & 0.4433 & 0.5502 & 0.5833 & 0.6338 & 0.0128 & 0.3662 & 0.8105 \\
    CDs → SAM → CWBJ & 0.4568 & 0.5642 & 0.6329 & 0.6056 & \textbf{0.0074} & 0.3944 & 0.7991 \\
    EffV2-UNet 1st Iter. & 0.4762 & 0.5875 & 0.6297 & 0.6531 & 0.0089 & 0.3469 & 0.8221 \\
    EffV2-UNet 2nd Iter. & \textbf{0.5155} & \textbf{0.6198} & 0.6360 & \textbf{0.6994} & 0.0096 & \textbf{0.3006} & \textbf{0.8449} \\
    \midrule
    \end{tabular}
\end{table*}
To evaluate SinkSAM's performance in sinkhole detection and segmentation, we designed an ablation experimental setup to determine the contribution of each component within the framework. For the ablation experiment study, EffV2-UNet trained on closed depressions (CDs) from the DAV2 dataset serves as the baseline. We tested four configurations reflecting key stages of  SinkSAM pipeline: (A) EffV2-UNet trained on CDs pseudo-labels vs. zero-shot SAM pseudo-labels prompted by the same CDs ($CDs \rightarrow SAM$); (B) EffV2-UNet trained on SAM pseudo-labels vs. SAM pseudo-labels generated using CWBJ and shape-aware filtering ($CDs \rightarrow SAM \rightarrow CWBJ$); (C) EffV2-UNet trained on SAM+CWBJ pseudo-labels vs. the self-trained EffV2-UNet, trained using its own predictions from a prior iteration ($CDs \rightarrow SAM \rightarrow CWBJ \rightarrow$ EffV2-UNet 1st Iter. $\rightarrow$ EffV2-UNet 2nd Iter.). All trained models, each using different pseudo-labels, were evaluated on the test dataset. See model examples in Figure~\ref{comparison_grid_models}.

In addition to the ablation experiments, we provide two benchmarking experiments on our dataset:
(1) We compared our self-supervised SinkSAM-Net with supervised learning approaches previously used in sinkhole segmentation studies. In this setup, an EffV2-UNet trained on human-labeled data serves as the baseline, allowing us to benchmark the performance of the self-supervised SinkSAM-Net framework (see Section~\ref{new_exp}).
(2) We compared monocular depth estimation (MDE) from DAV2 with the photogrammetric DEM available for part of our dataset ($n = 216$), providing a baseline for the quality of the estimated depth data.

\begin{table*}[ht!]
    \centering
    \caption{Performance of Models Trained on Varying Amounts of Human-Labeled Data}
    \label{tab:human-labels-variation}
    \footnotesize
    \begin{tabular}{p{3.5cm} >{\centering\arraybackslash}p{2 cm} >{\centering\arraybackslash}p{1.5cm} >{\centering\arraybackslash}p{1.5cm} >{\centering\arraybackslash}p{1.5cm} >{\centering\arraybackslash}p{1.2cm} >{\centering\arraybackslash}p{1.2cm} >{\centering\arraybackslash}p{2cm}}
    \toprule
    \textbf{Train Dataset} & \textbf{Sinkhole IoU ↑} & \textbf{F1 Score ↑} & \textbf{Precision ↑} & \textbf{Recall ↑} & \textbf{FPR ↓} & \textbf{FNR ↓} & \textbf{Balanced Acc. ↑} \\
    \midrule
    Human Labels data (N=450) & 0.4803 & 0.5757 & 0.6838 & 0.5869 & 0.0054 & 0.4131 & 0.7907 \\
    Human Labels data (N=900) & 0.5222 & 0.6199 & \textbf{0.7756} & 0.5857 & \textbf{0.0032} & 0.4143 & 0.7912 \\
    Human Labels data (N=1800) & \textbf{0.5508} & \textbf{0.6517} & 0.7438 & \textbf{0.6501} & 0.0041 & \textbf{0.3499} & \textbf{0.8230} \\
    \bottomrule
    \end{tabular}
\end{table*}

\subsection{Depth Anything V2 vs Photogrammetric DEM}

In this comparison, no model training was performed; instead, we directly applied the "fill sinks" algorithm to the computed depth maps to yield segmentation results. From this test, we clearly observed that DAV2 significantly outperformed the photogrammetric DEM as input data for delineating closed depressions. The photogrammetric DEM appeared too smooth, with lower differences between grid-cell values in the immediate vicinity of the depressions, compared with DAV2. Depressions delineated by DAV2 achieved an F1-score of 0.5234, while those based on the photogrammetric DEM achieved only 0.2075.

\subsection{CDs pseudo-labels vs SAM pseudo-labels ($CDs \rightarrow SAM$)}

In this comparison, we assess the central concept of SinkSAM-Net: the contribution of SAM to the refinement of delineations based on closed depressions derived from DAV2 using the "fill sinks" algorithm. The results indicate that SAM indeed refines and improves the training dataset, which is ultimately reflected in the segmentation performance of EffV2-UNet. In terms of pixel-level metrics, EffV2-UNet trained on SAM-prompted masks achieves an F1-Score of 0.55 compared with 0.54 for models trained on closed depressions, and an IoU of 0.44 compared with 0.43. SAM reduces the rate of incorrectly segmented pixels, resulting in a lower FNR and higher balanced accuracy. This improvement in metrics indicates that SAM, when prompted by these bounding boxes, successfully refines the closed depression extraction process using RGB sinkhole representations.

\subsection{SAM pseudo-labels ($CDs \rightarrow SAM$) vs SAM + CWBJ pseudo-labels ($CDs \rightarrow SAM \rightarrow CWBJ$)}

This section presents an evaluation of coordinate-wise bounding box jittering (CWBJ) and mean-based shape filtering. The results demonstrate that CWBJ and the filtering mechanism provide a more reliable training dataset for EffV2-UNet. For pixel-level segmentation, EffV2-UNet trained on this improved dataset achieves an F1-Score of 0.56, compared with 0.55 for the model trained using pseudo-labels generated from a single SAM-prompted bounding box. Similarly, IoU increases from 0.44 to 0.46. The single-box prompting approach tends to overestimate sinkhole extent, leading to more background segmentation and a higher false positive rate (FPR).

\subsection{SAM + CWBJ pseudo-labels ($CDs \rightarrow SAM \rightarrow CWBJ$) vs Self-Trained EffV2-UNet pseudo-labels}

EffV2-UNet was trained using pseudo-labels generated by a previous EffV2-UNet trained with SAM + CWBJ prompted by CDs ($CDs \rightarrow SAM \rightarrow CWBJ$), effectively relabeling the entire training dataset. Passing the dataset through SAM + CWBJ indeed improved the training quality. EffV2-UNet trained with these pseudo-labels achieved an F1-score of 0.59, compared with 0.57 for the model trained on original SAM + CWBJ pseudo-labels, and an IoU of 0.48 compared with 0.46. A second iteration of relabeling and retraining further improved the performance, achieving an F1-score of 0.62 and an IoU of 0.52, representing the best performance obtained with the self-supervised SinkSAM-Net framework. Notably, after the second iteration of EffV2-UNet training, we applied CWBJ and SAM on the results, improving slightly just F1-score from 0.6198 to 0.6231. This indicates proximity to reaching self-improving limit.

\subsection{Self-supervised vs Supervised}

Finally, we compared the best results of self-supervised SinkSAM-Net with identical target EffV2-UNet model trained on human-labeled data. The target model was trained using three configurations: full training set; half of the dataset; and one-quarter of the dataset. As expected, the performance of the supervised model increased with the amount of labeled data, achieving its best results on the full dataset of 1800 sinkholes, with an F1-Score of 0.6517 and an IoU of 0.5508. These represent the theoretical upper bound for our dataset.

The best SinkSAM-Net configuration, based entirely on self-supervised pseudo-labeling, slightly trails this upper bound, with an F1-Score of 0.6198 and an IoU of 0.5155, corresponding to approximately 95\% and 93\% of the fully supervised model’s performance, respectively. Interestingly, when the supervised model was trained on only half of the dataset (900 samples), it slightly outperformed the self-supervised SinkSAM-Net, with an F1-Score of 0.6199 and an IoU of 0.5222. However, the supervised model trained on just one-quarter of the data (450 samples) achieved an F1-Score of 0.5757 and an IoU of 0.4803, which is notably lower than the self-supervised SinkSAM-Net.

\begin{figure}[ht!]
    \centering
    \includegraphics[width=\linewidth]{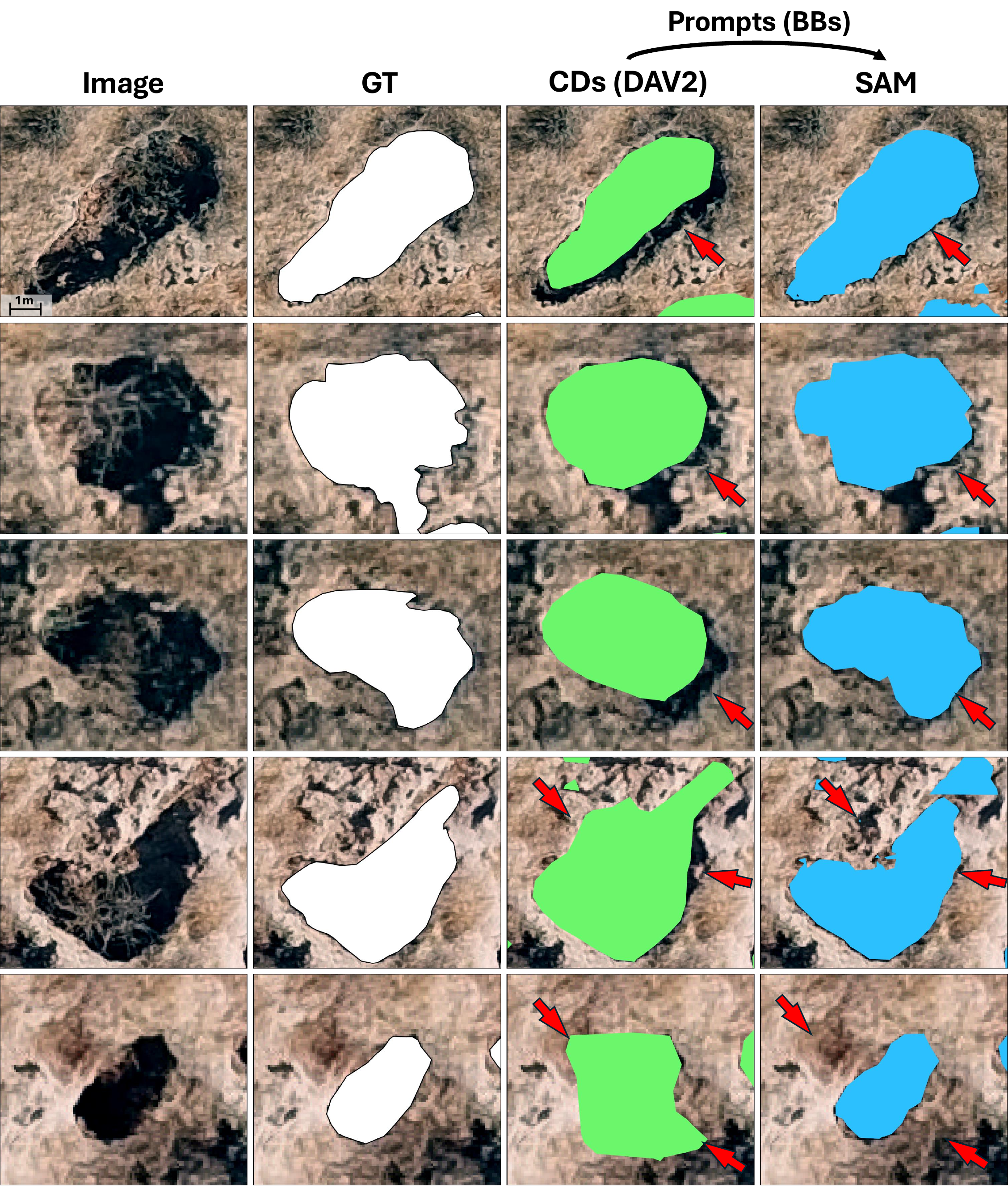}
    \caption{SAM prompted by closed depressions: SAM improved sinkhole segmentation, although closed depressions (DAV2) tend to either overestimate or underestimate sinkhole areas, leading to inaccurate bounding boxes. SAM (zero-shot), especially when prompted with smaller boxes, effectively refines the output, providing more accurate delineation of actual sinkholes (CDs - Closed Depressions).}
    \label{Prediction_test_set}
\end{figure}

\begin{figure*}[ht!]
   \centering
    \includegraphics[width=1\linewidth]{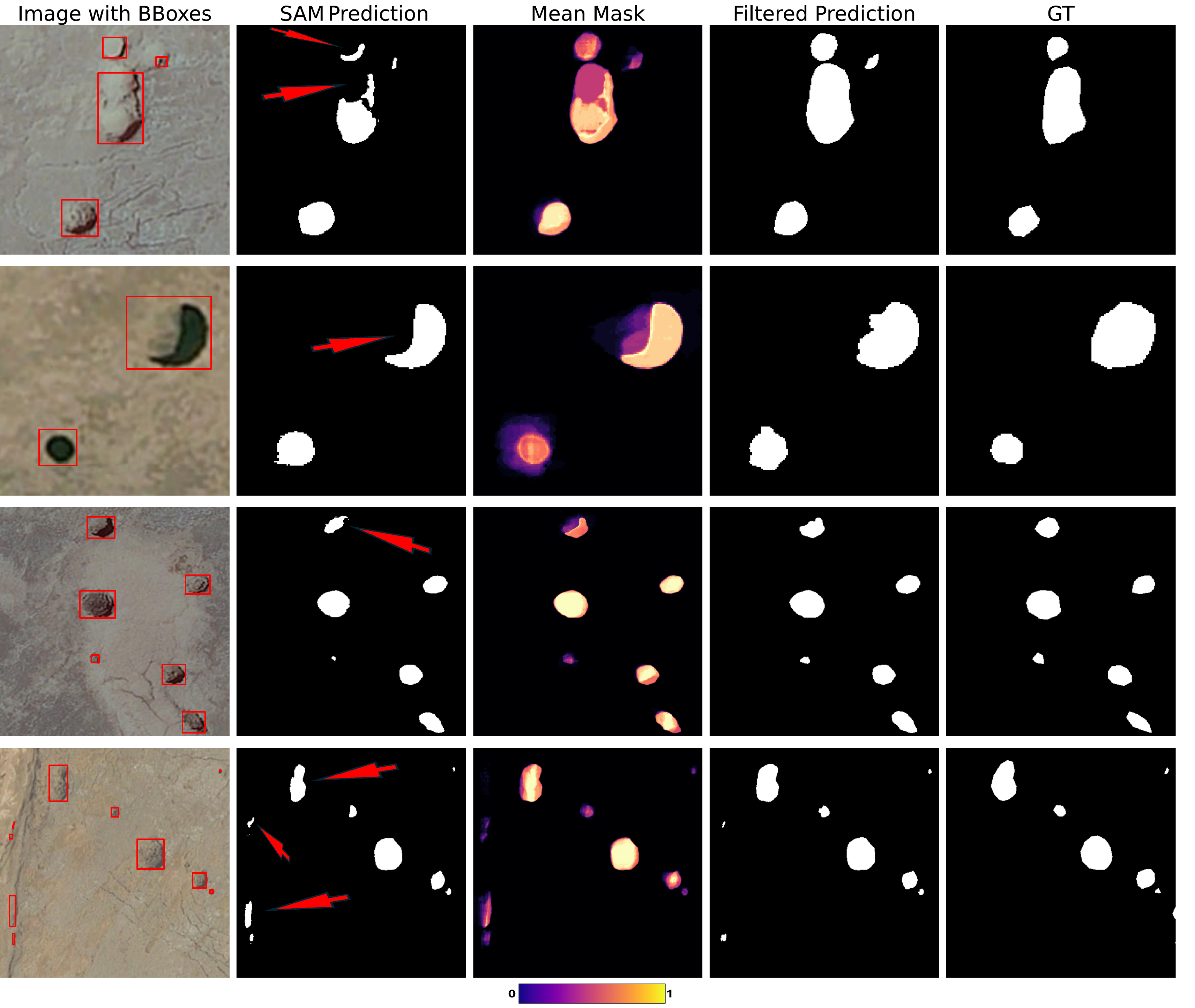}
    \caption{Mean Base Shape-Aware Filtering. A single SAM prediction using a CD-derived bounding box may under-segment or over-segment sinkholes due to inaccurate bounding boxes. By applying coordinate-wise jittering and Monte Carlo simulations, followed by thresholding the averaged mask, boundary delineation is substantially improved. Non-circular artifacts are further filtered using Polsby–Popper circularity index, yielding more accurate and reliable pseudo-labels.}
    \label{cwbj}
\end{figure*}
\section{Discussion}
The series of experiments conducted here highlight three key features of the self-supervised SinkSAM-Net framework: (1) automatic refinement of closed depressions via pixel-level RGB image segmentation; (2) integration of monocular depth for enhanced sinkhole segmentation; and (3) knowledge transfer from the heavy, parameter-rich SAM model into a lightweight, end-to-end, prompt-free semantic segmentation architecture. This architecture incorporates an EfficientNetV2-based U-Net with a low parameter count and minimal computational requirements, thereby enhancing the practical applicability of the framework.

Throughout this work, we emphasize SAM's potential to extract refined, high-quality pseudo-labels using weak prior domain knowledge. This prompt-based strategy forms the foundation of our framework. For the first time, it enables a high-performance self-supervised segmentation approach that generalizes across a diverse sinkhole database, capturing a wide range of common sinkhole geometries.
When compared with a model trained on human-labeled data, the results indicate that the self-supervised SinkSAM-Net slightly underperformed in terms of F1 score and IoU on the full dataset, achieving an F1 score of 0.6198 compared with 0.6517 for the supervised model. However, when trained on only half of the data, both approaches show comparable performance in terms of F1 score and IoU (Table~\ref{tab:full-metrics-comparison} and Table~\ref{tab:human-labels-variation}).
Notably, the self-supervised framework has the potential to scale across large datasets without any human intervention, allowing it to process more data and approach high-performance levels without the limitations imposed by manual annotation or human resource constraints.

SinkSAM-Net can serve as a strong starting point and a solid benchmark for future self-supervised sinkhole segmentation frameworks. This work also represents an important step toward establishing a foundation model for sinkhole detection, overcoming the limitations of previous studies, which were often constrained to specific regions or seasons with limited diversity. We also note several key features that further support the framework robustness and generalizability.

\subsection{Refinement of Closed Depressions using SAM}

The DEM-based sinkhole delineation method, "fill sinks," has become widely utilized in topographic analyses and geomorphometric applications. Its integration into ArcGIS Pro made it accessible to a broad range of researchers, engineers, and soil conservationists for various purposes \citep{PARDOIGUZQUIZA2021480,HOFIERKA2018265,article13}. As discussed in the \textit{Related Work} section, relying solely on DEM-based sinkhole delineation, without RGB support, introduces two main challenges: (1) false detection of non-sinkhole features \citep{HOFIERKA2018265}; and (2) inaccurate segmentation of sinkhole boundaries \citep{LINDSAY20061192}. 

Consistent with these limitations, our results demonstrate that closed depressions delineated using "fill sinks" tend to either under- or overestimate sinkhole boundaries. This results in area coverage discrepancies and incorrectly segmented pixels. In contrast, SinkSAM-Net integrates computed closed depressions from 3D data as bounding boxes to prompt SAM for pixel-level RGB segmentation. This improves sinkhole detection and delineation compared with traditional DEM-based methods (see Fig.~\ref{Prediction_test_set}). SinkSAM-Net enables automatic refinement of sinkhole boundaries using RGB data, thereby achieving a level of accuracy previously obtained only through manual post-processing by human annotators \citep{10943264,chen2015semi}.

Zero-shot SAM was applied in RS studies with promising results \citep{osco2023segmentmodelsamremote}. For instance, it was used to segment glaciers and geological features in a zero-shot setting \citep{GIANNAKIS2024115797,shankar2023semantic}. However, Zero-shot SAM is prone to over-segmentation or under-segmentation, especially when object boundaries are amorphous and therefore unclear \citep{ji2024segment}. This is a methodological challenge because many objects on the earth surface are amorphous and hard to segment from RS data. \citep{10613866,osco2023segmentmodelsamremote}. 
In line with these findings on different objects, SAM’s performance in sinkhole segmentation here is reasonable in zero-shot mode, but, it tended to segment parts of the background, such as shadows and shrubs, leading to a higher false positive rate, especially when object boundaries were fuzzy.

Specifically, applying zero-shot SAM improved sinkhole segmentation: the IoU increased from 0.4318 for the “fill sinks” method to 0.4433 for SAM. This improvement holds for both large bounding boxes (which include sinkholes and background) and smaller boxes that partially capture sinkholes (Fig.~\ref{Prediction_test_set}). However, when SAM was prompted by non-sinkhole features, the model did not always filter them out. In cases prompts involving non-sinkhole, SAM occasionally produced erroneous segmentation \citep{China}.

To address the zero-shot challenge of sinkhole segmentation, we introduced prompt augmentation and mean segmentation analysis based on Monte Carlo simulations with SAM, followed by shape-based filtering. The framework begins with learning from closed depressions (CDs) as pseudo-labels and eventually evolves into a CD prompt-free pipeline, reducing dependence on imperfect topographic delineations.
Inspired by recent developments in medical imaging, we augmented bounding box prompts by aggregating multiple predictions generated through random perturbations \citep{10.1007/978-3-031-47425-5_33}. To adapt this strategy to our remote sensing (RS) sinkhole segmentation task, we introduced two key modifications:
(1) Unlike the prompt-augmentation techniques used in the medical-imaging domain, where bounding-box augmentation is typically applied when a single object appears in each image \citep{zhou2024medsamuuncertaintyguidedautomultiprompt}, remote-sensing (RS) imagery often contains multiple objects within a single scene. Therefore, we applied prompt augmentation independently to each bounding box, treating each region as a separate geometric entity. This enabled the framework to filter out false positives at the region level.
(2) In contrast to the slightly inaccurate human-annotated prompts commonly used in medical imaging \citep{zhu2024medicalsam2segment}, the automatically derived closed depression (CD) prompts in RS do not fully enclose actual sinkhole geometries and may only partially cover them, leading to more severe prompt inaccuracies \citep{brazil}. Thus, instead of merely shifting the center of the bounding boxes \citep{10.1007/978-3-031-47425-5_33}, we jittered each corner coordinate independently. This approach allowed the prompts to explore uncertain boundaries, particularly when the RGB appearance of sinkholes deviated from the initial CD-derived bounding box.

This approach is especially valuable in RS tasks with inaccurate pseudo-labels, for two key reasons. \textit{First}, small prompt perturbations allow SAM to segment even when the bounding box is imperfect. \textit{Second}, In the absence of an object, SAM’s unstable and noisy outputs produce a low-confidence mean mask, thereby effectively ignoring false positives.
Finally, we applied CWBJ in conjunction with the Polsby–Popper (PP) index, leveraging known circular geometry of sinkholes \citep{RAFAELI2023107511}. In most cases, valid sinkholes yielded PP values greater than 0.3. This integration of prior knowledge further reduced false positives by eliminating non-circular detections. This improvement, which takes the system from single-prompt SAM to a mean-based, shape-aware filtering approach, can be visually inspected in Figure ~\ref{cwbj}.

\subsection{Monocular Depth Estimation}

Depth information may offer an essential data source for accurate sinkhole segmentation and detection. Yet, high-resolution DEMs are not always readily available \citep{hummel2011comparison}. LiDAR, while providing high-resolution and accurate elevation data, is rarely used at the local scale due to its high cost. Conversely, low resolution LiDARs are not usable for sinkhole segmentation. For instance, a 1 m airborne LiDAR DEM was found inadequate for detecting small soil piping-related sinkholes \citep{BERNATEKJAKIEL2021107591}. Photogrammetric DEMs are more accessible, but often suffer inaccuracies and misalignment with orthomosaic, which can lead to an under representation of sinkholes, making them less reliable for segmentation or even creating accurate topographic models \citep{UYSAL2015539}.

\begin{figure}[ht!]
    \centering
    \includegraphics[width=\linewidth]{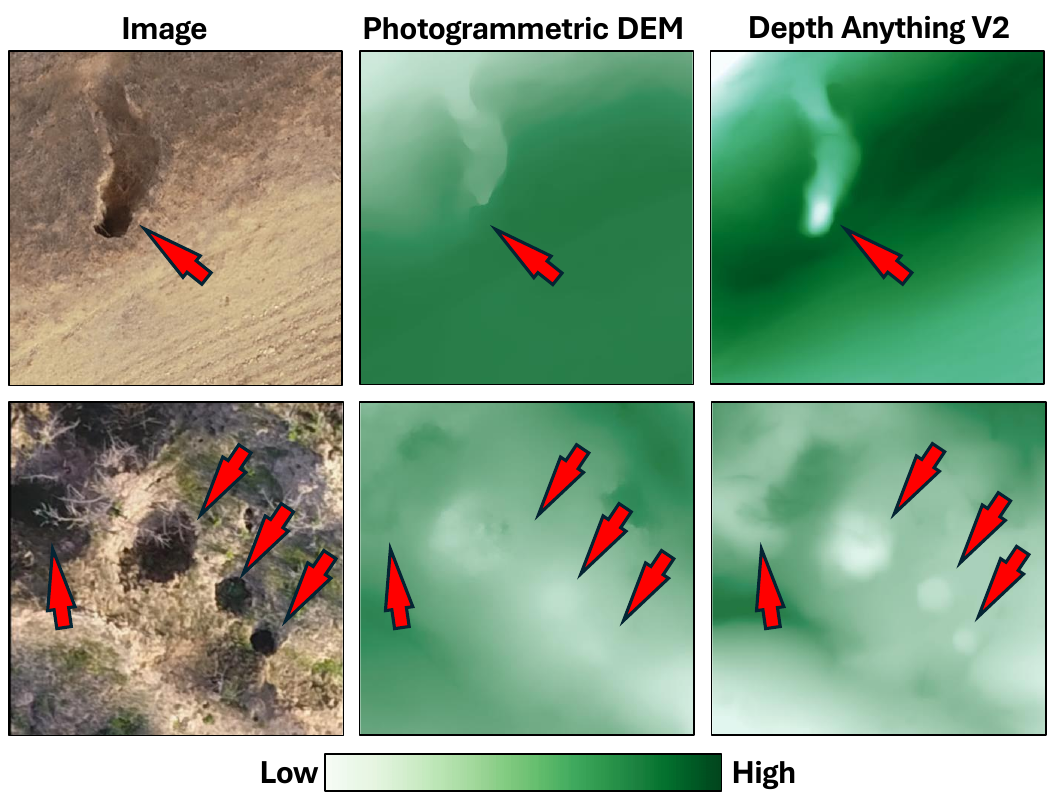}
    \caption{\small Comparison of sinkhole delineation using photogrammetric DEM and DAV2. DAV2 more accurately captures sinkhole boundaries and fully delineates all sinkholes, while the DEM-based method only partially fills depressions, often missing smaller sinkholes.}
    \label{DEMvsMONOD}
\end{figure}

Similar to previous studies, we observed inaccuracies in sinkhole detection, encountered with photogrammetric DEMs. To improve model accuracy, we used MDE (Monocular Depth Estimation) foundation models, which have gained traction in depth estimation tasks by enabling zero-shot or fine-tuning predictions \citep{birkl2023midasv31model}. Notably, state-of-the-art DAV2 model has delivered impressive results across multiple datasets, including aerial imagery, outperforming equivalent foundation models \citep{yang2024depthv2}. Since its introduction, DAV2 was applied  in agricultural canopy height measurements \citep{cambrin2024depth}, offering a highly efficient and high-performing solution that surpasses current methods with superior or comparable accuracy.

Our findings demonstrate that DAV2, excels in depth estimation using a single RGB drone image as an input. DAV2 allowed to detect fine sinkholes missed by photogrammetric DEMs, and provided more precise sinkhole boundaries delineation (Fig. \ref{DEMvsMONOD}). DAV2’s depth maps are aligned geometrically with an RGB image, eliminating small offsets typically found in photogrammetric DEMs and ultimately generating more accurate prompts for SAM.

The current paper marks the first application of DAV2, on RS drone imagery, for detecting geomorphic features of interest. Specifically, it is the first time a raster-based algorithm has been applied to monocular depth estimation for geomorphic feature extraction, a process that traditionally done using LiDAR or photogrammetric DEMs \citep{ZUMPANO2019213}. DAV2 can be further fine-tuned on drone imagery from sinkhole regions for even more accurate depth estimation, and it can also be switched to metric mode, enabling the extraction of both the geometric dimensions and the boundaries of sinkholes.

\subsection{Target EffV2-UNet model}
Despite success in a wide range of applications, the original Segment Anything Model (SAM), particularly SAM with the large version of ViT encoder, suffers significant limitations due to slow runtime and high computational cost \citep{zhang2024efficientvit}. These challenges become even more pronounced when SAM is deployed in resource-constrained or real-time environments, such as edge devices and mobile applications \citep{wang2024repvitsamrealtimesegmenting}.

SAM 2.1 also adopts the pre-trained Hiera \citep{ryali2023hiera} as its image encoder, which is more efficient than ViT-l, with expectations of higher speed. In SAM 2.1-L, the Hiera-L image encoder contains approximately 226M parameters, while the total model size is 235M \citep{ravi2024sam}. Depth Anything V2 is also based on a large ViT encoder with a total of 335.3M parameters \citep{yang2024depthv2}. Compared to ViT-based models, the EfficientNetV2 family of convolutional networks offers faster training speeds and better parameter efficiency, while achieving comparable accuracy \citep{tan2021efficientnetv2smallermodelsfaster}. Our EffV2-UNet has only 6M parameters and processes more then 20 images per second compared with SAM 2.1 that processes only three images.

Thus, the entire process of generating pseudo-labels is highly demanding in terms of computational resources. In addition to the two high-parameter foundation models, the CWBJ approach increases time complexity due to the multiplication of Monte Carlo simulations per bounding box. Consequently, our motivation was to ultimately train a lightweight CNN model that could be deployed in real-world scenarios using mobile devices with limited processing power in the field. Once trained on a large dataset, the lightweight segmentation model provides fast inference times.

Furthermore, this approach transforms the entire model into an end-to-end, learning-based, prompt-free framework, reducing dependency on artifacts introduced by monocular depth estimation (MDE), which may lead to inaccurate prompts. For example, during depth estimation, we observed that Depth Anything V2 may produce finely detailed but inverted depth maps, particularly in low-quality aerial images with poor lighting conditions. In contrast, once the CNN learns the RGB appearance of sinkholes, it becomes more stable and reliable than MDE CDs computations. During training, the model is more tolerant of noise: some training samples may be slightly inaccurate or even misleading, but the majority still guide the feature maps to effectively capture sinkholes \citep{Veit_2017_CVPR}. However, during inference, incorrect prompts can lead to significant segmentation errors. To mitigate this risk across diverse RGB image sources, we propose a fully data-driven approach based solely on RGB imagery, offering greater consistency, generalizability, and performance.

\section{Conclusion}
 
We present SinkSAM-Net, an innovative automated system built upon the Segment Anything Model that detect and segments soil sinkholes from individual RGB images, through self-supervised learning techniques. The system combines "fill sinks" calculations with SAM's prompt-driven segmentation capabilities, successfully processing areas hidden by plant growth and shadows. SinkSAM-Net approach achieves exceptional performance, comparable to expert human-supervised benchmarks on the evaluation dataset. These results position SinkSAM-Net as a generalized and accessible framework for sinkhole segmentation. The SinkSAM-Net framework may be applied to agricultural land management, geological research, hazard mapping and risk assessment, and early warning systems. It can be improved in four key areas:
(1) enhancing prompt generation from depth information by employing more advanced computational methods or topographic indices combined with classifiers, to reduce overestimation and generate more reliable prompts for SinkSAM-Net;
(2) implementing SinkSAM-Net with LiDAR data to evaluate its performance in comparison with the DAV2-based approach;
(3) integrating drone-acquired LiDAR data to fine-tune DAV2 on sinkholes images for improved depth estimation capabilities from individual RGB images; 
and
(4) enhancing the sinkhole database by incorporating examples from more humid environments characterized by thicker vegetation coverage.

\section*{Acknowledgment}

We thank the Ministry of Agriculture Chief Scientist, grant number 16-17-0005, 2022, and the Negev Scholarship by the Kreitman School of Ben-Gurion University of the Negev for supporting Osher Rafaeli’s PhD studies.
We also thank the reviewers and editor for constructive comments and suggestions during the revision process, which greatly improved the manuscript.

\bibliographystyle{elsarticle-harv} 
\bibliography{cas-refs}





\end{document}